\documentclass{article}

\usepackage{arxiv}

\usepackage[utf8]{inputenc} % allow utf-8 input
\usepackage[T1]{fontenc}    % use 8-bit T1 fonts
\usepackage{hyperref}       % hyperlinks
\usepackage{url}            % simple URL typesetting
\usepackage{booktabs}       % professional-quality tables
\usepackage{amsfonts}       % blackboard math symbols
\usepackage{microtype}      % microtypography
\usepackage{lipsum}
\usepackage{graphicx}
\graphicspath{ {./images/} }

\title{Combining Generative Artificial Intelligence (AI) and the Internet: Heading towards Evolution or Degradation?}

\author{
 Gonzalo Mart\'inez \\
  Universidad Carlos III de Madrid\\
  28911 Madrid, Spain \\
  \texttt{gonzmart@pa.uc3m.es} \\
  \And  
 Lauren Watson\\
  School of Informatics \\
  University of Edinburgh \\
  \texttt{lauren.watson@ed.ac.uk}
  \And
 Pedro Reviriego \\
  Universidad Polit\'ecnica de Madrid\\
  28040 Madrid, Spain \\
  \texttt{pedro.reviriego@upm.es} \\
  \And
  Jos\'e Alberto Hern\'andez\\
  Universidad Carlos III de Madrid\\
  28911 Madrid, Spain \\
  \texttt{jahgutie@it.uc3m.es} \\
  \And
  Marc Juarez\\
  School of Informatics \\
  University of Edinburgh \\
  \texttt{mjuarez@inf.ed.ac.uk}
  \And
  Rik Sarkar\\
  School of Informatics \\
  University of Edinburgh \\
  \texttt{rsarkar@inf.ed.ac.uk}
}

\begin{document}
\maketitle
\begin{abstract}

In the span of a few months, generative Artificial Intelligence (AI) tools that can generate realistic images or text have taken the Internet by storm, making them one of the technologies with fastest adoption ever. Some of these generative AI tools such as DALL-E, MidJourney, or ChatGPT have gained wide public notoriety. Interestingly, these tools are possible because of the massive amount of data (text and images) %available on the Internet. The tools are trained on massive data sets that are 
scraped from Internet sites. Now, these generative AI tools are creating massive amounts of new data that are being fed into the Internet. Therefore, future versions of generative AI tools will be trained with Internet data that is a mix of original and AI-generated data. As time goes on, increasing volumes of data generated by different versions of AI will populate the Internet. 

This raises a few intriguing questions: how will future versions of generative AI tools behave when trained on a mixture of real and AI generated data? Will they evolve and improve with the new data sets or degenerate? Will evolution introduce biases in subsequent generations of generative AI tools?  In this document, we explore these questions and report some initial simulation results using a simple image-generation AI tool. These results suggest that the quality of the generated images degrades as more AI-generated data is used for training thus suggesting that generative AI may degenerate. Although these results are preliminary and cannot be generalised without further study, they serve to illustrate the potential issues of the interaction between generative AI and the Internet. 
%The document also discusses how natural intelligence evolves differently from generative AI and proposes future work to better understand the impact of training generative AI with AI-generated data. 

\end{abstract}

% keywords can be removed

\section{Introduction}
\label{sec:introduction}

Traditional applications of Artificial Intelligence (AI) have focused on the detection or classification of objects, for example detecting pedestrians in the images captured by in-vehicle cameras \cite{PedestrianCNN} or classifying the results of a medical test from x-ray images \cite{BreastCancerCNN}. More recently, AI models that can generate images, text, or even videos and 3D objects have been developed and made publicly available \cite{GenerativeAIModels}. These models have been widely adopted and some of them have now millions of users, like for example ChatGPT for text or DALL-E for image generation. 

In the case of image generation, AI models such as DALL-E, MidJourney, and Stable Diffusion \cite{StableDiffusionMidJourneyDALLE} are now widely used to generate many different types of images with applications in fields as diverse as art \cite{AIART}, architecture \cite{AIArchitecture} or cultural heritage \cite{AIMosaics} among other. These AI tools are also used to generate illustrations and content for websites and magazines and also by end-users that upload the images to social networks and other applications. These generative AI models such as DALL-E \cite{DALLE} or CogView \cite{CogView} are trained on large data sets of images that in most cases are obtained from the Internet by scraping websites \cite{LAIN400M,LAION5B}. 

As AI-generated content populates the Internet, AI-generated images will be scrapped and may end up in future training data sets. This poses the following question: \emph{what is the effect of this feedback loop on AI tools that use AI-generated images? \cite{Datasetcorruption}} An initial study has shown that current AI-generated images tend to degrade the performance of AI image tools \cite{Datasetcorruption}. In this paper, we study the long-term effects that AI image generation tools can have. For example, we investigate whether repeatedly feeding AI generative models with AI-generated data over time leads to \emph{degeneration}, so newer models producing lower quality results as they are trained with more and more AI-generated samples. Our initial experiments on a simple diffusion model show that degeneration is indeed possible.
We also discuss other potential issues as generative AI and the Internet feed each other over time and propose future work and experiments to better understand the impact of this closed loop interaction between generative AI and the Internet\footnote{In fact, this paper is a first work in progress version and we are currently conducting additional experiments to produce a second more complete version.}.

%and also compare the generative AI models' evolution with that of natural intelligence, showing that there are fundamental differences that make natural intelligence more robust to this type of degeneration. Based on that analysis, we propose future work and experiments to better understand the impact of this closed loop interaction between generative AI and the Internet .

\section{Initial Evaluation of Generative AI evolution}
\label{sec:evaluation}

\begin{figure}[h]   
    \centering
    \includegraphics[scale=0.6]{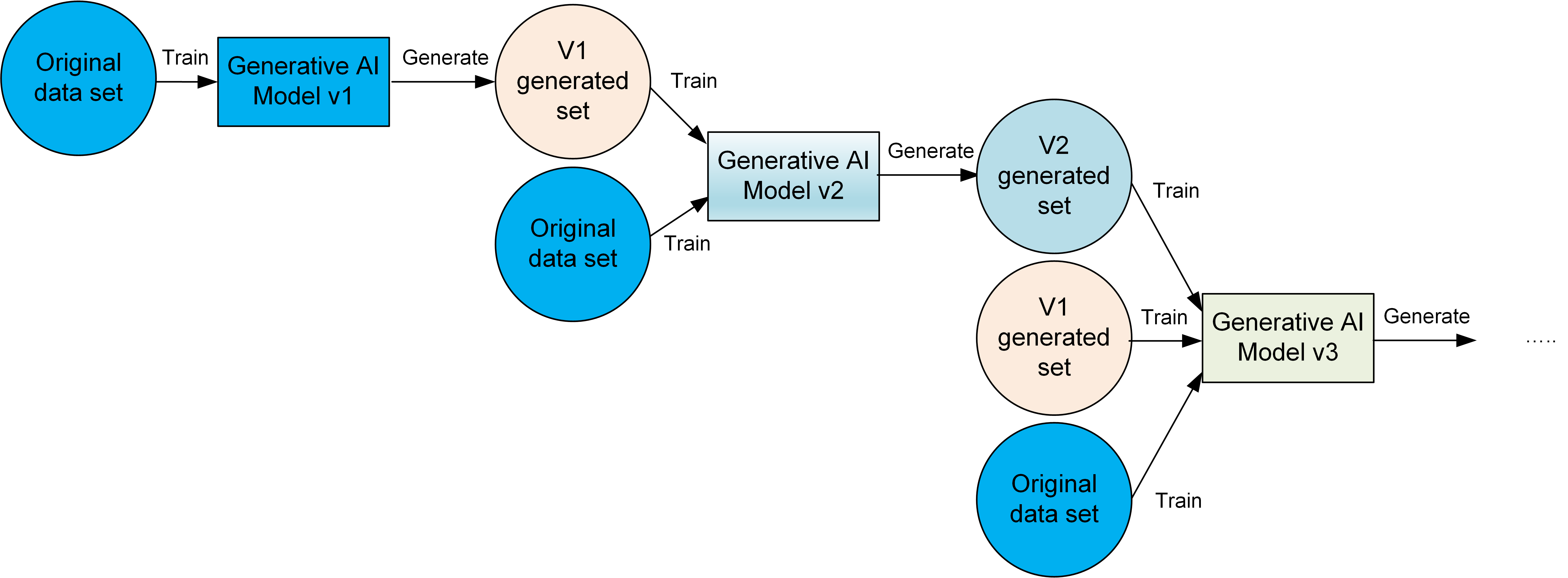}
    \caption{\textbf{Simulation model for the evolution of generative AI}}
    \label{Fig:AI_Evolution}
\end{figure}

To study the evolution of generative AI, we consider the simulation model in Figure~\ref{Fig:AI_Evolution} in which the original data set is augmented with AI-generated images and then trained to generate a new version of the AI-generation model. Then, the new model generates images that are added to the data set to train yet another model, the third-generation model. This process may continue for $n$ generations. This simple model tries to capture what could happen on the Internet as AI-generated images are incorporated to the training datasets of new models. 

\begin{figure}
    \centering
    \includegraphics[scale=0.34]{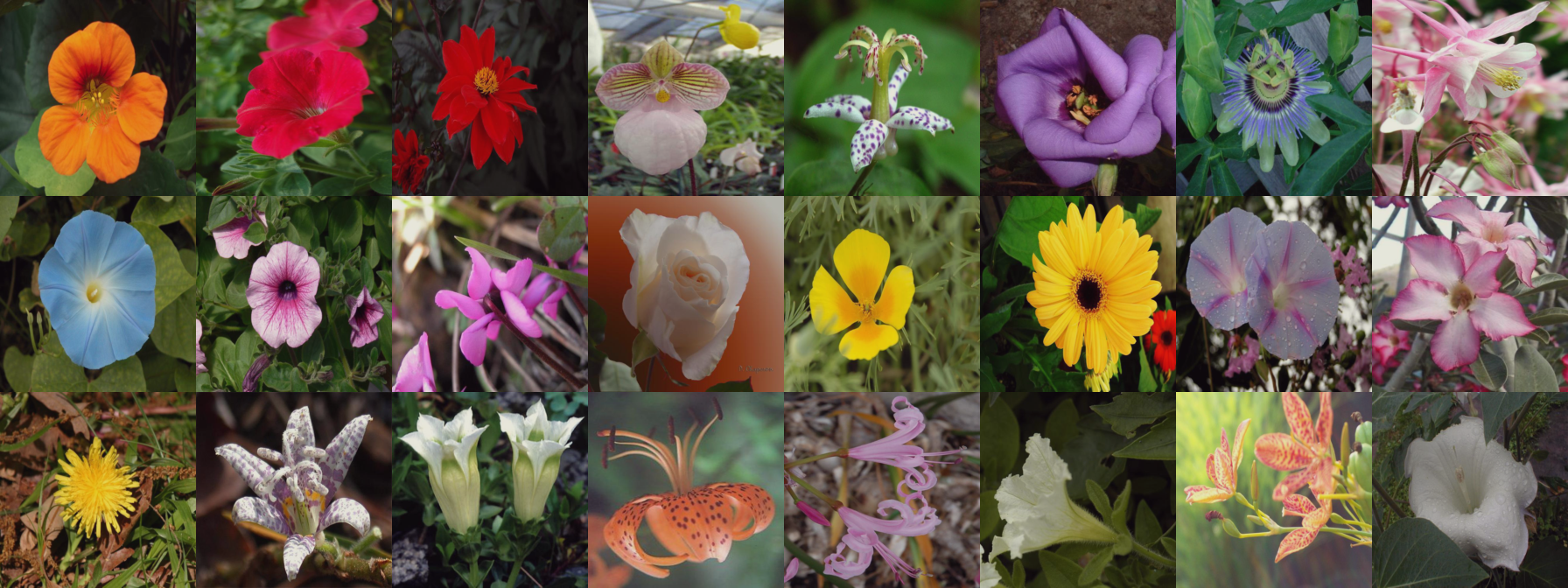}
    \caption{\textbf{Examples of real flowers, Original dataset}}
        \label{Fig:FlowersOriginal}
\end{figure}

\begin{figure}
    \centering
    \includegraphics[scale=0.6]{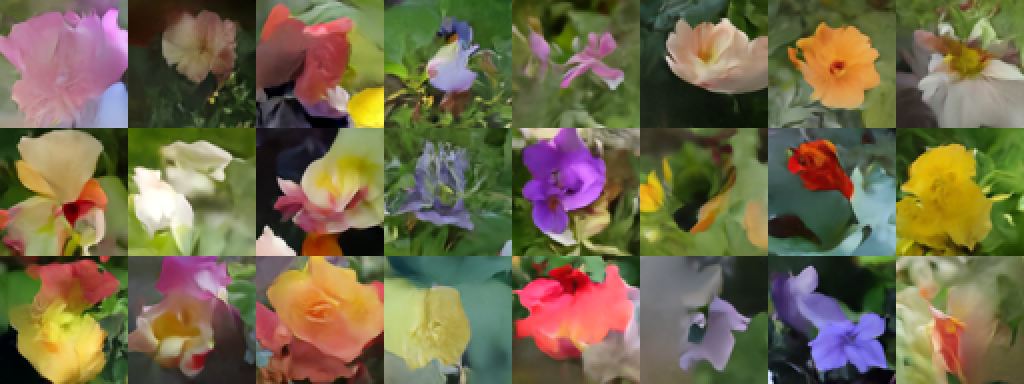}
    \caption{\textbf{Examples of synthetic flowers, $V_1$ generated set}}
    \label{Fig:FlowersAIv1}
\end{figure}

To test the model in Figure~\ref{Fig:AI_Evolution} for generative AI, we use a Denoising Diffusion Implicit Model (DDIM)\footnote{See A. Beres: "Denoising Diffusion Implicit Models", available at \url{https://keras.io/examples/generative/ddim/}, last access February 2023.}. For training the model, we use the Oxford Flowers dataset, consisting of 1023 images of flowers classified into 102 categories\footnote{See M.-E. Nilsback, A. Zisserman, "102 Category Flower Dataset", available at \url{https://www.robots.ox.ac.uk/~vgg/data/flowers/102/}, last access February 2023.}. We train a Generative AI model with the Oxford Flowers dataset and 40 epochs. Figure~\ref{Fig:FlowersOriginal} shows a few examples of the original images used for training and Figure~\ref{Fig:FlowersAIv1} shows the images generated by the diffusion model. These synthetic flowers comprise the first $V_1$ generated dataset, as shown in Figure~\ref{Fig:AI_Evolution}. 

In the next experiments, we mix both real data and AI generated data ($V_1$) to re-train the diffusion model and create a second generation of synthetic flowers ($V_2$). Let $\alpha$ denote the relative number of synthetic data elements generated with the latest AI version (flowers in this case) included during the training process compared to the size of the existing training set to obtain the next offspring. Thus, $\alpha$ controls the amount of AI generated elements introduced for the next generation. For example, when $\alpha=1$, at each iteration, the number of elements added to the training set generated with the latest version of the AI tool equals the total number of elements used to train the previous generation. 

\begin{figure}
    \centering
    \includegraphics[scale=0.6]{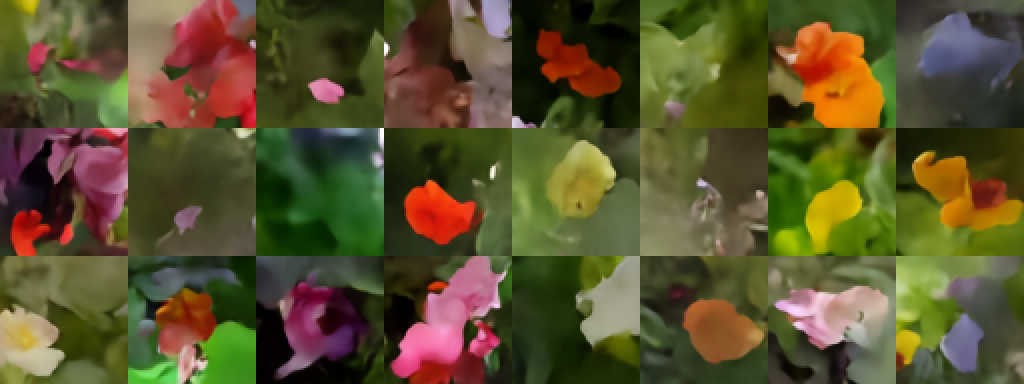}
    \includegraphics[scale=0.6]{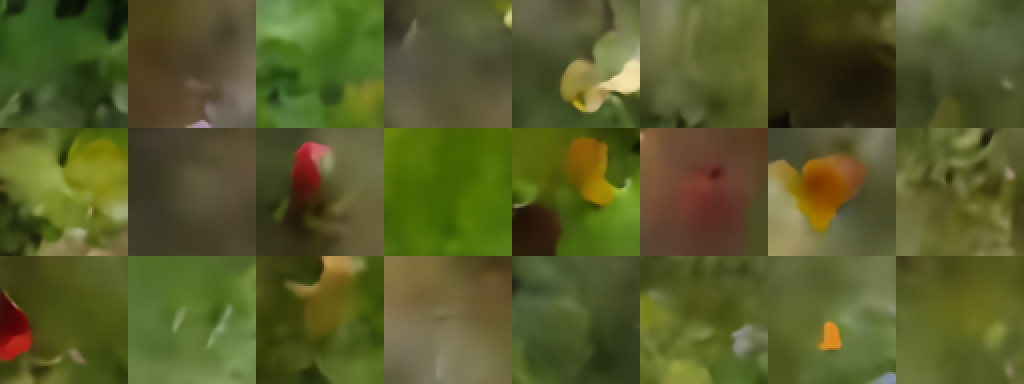}
    \includegraphics[scale=0.6]{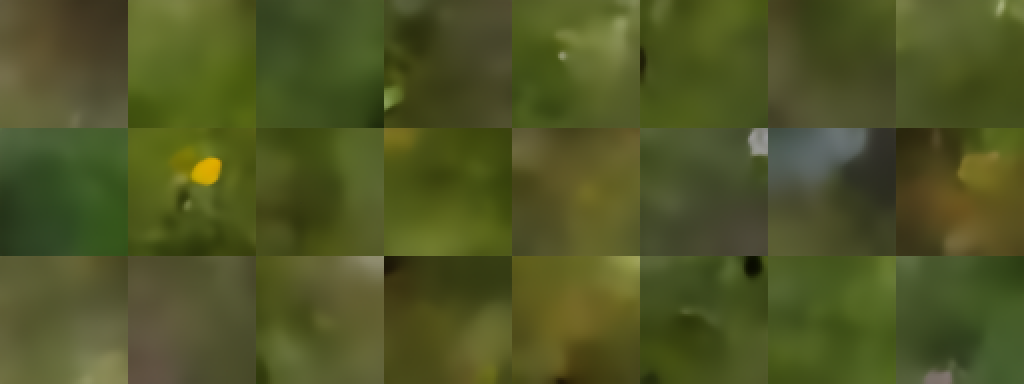}
    \caption{\textbf{Examples of images generated by the second $V_2$ (top), third $V_3$ (middle), and fourth $V_4$ (bottom) diffusion models trained with the original images and subsequent synthetic versions ($\alpha=1$)}}
    \label{Fig:FlowersAIv234a1}
\end{figure}

Figure~\ref{Fig:FlowersAIv234a1} shows examples of second, third, and fourth-generation synthetic flowers for $\alpha=1$, that is, at each generation $V_n$, the diffusion model is trained with as many synthetic flowers $V_{n-1}$ as original flowers and flowers generated with previous AI models. It can be observed that the quality of the images quickly degrades at each new iteration and the fourth AI model is unable to generate any flower with sufficient quality.

\begin{figure}
    \centering
    \includegraphics[scale=0.6]{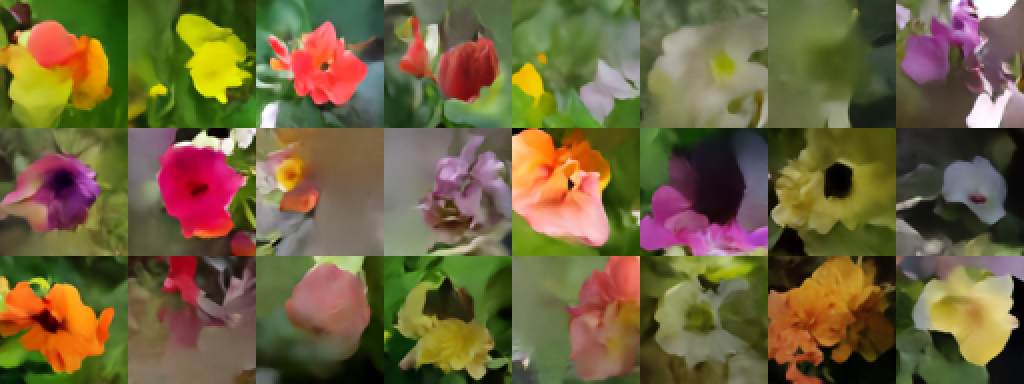}
    \\
    \includegraphics[scale=0.6]{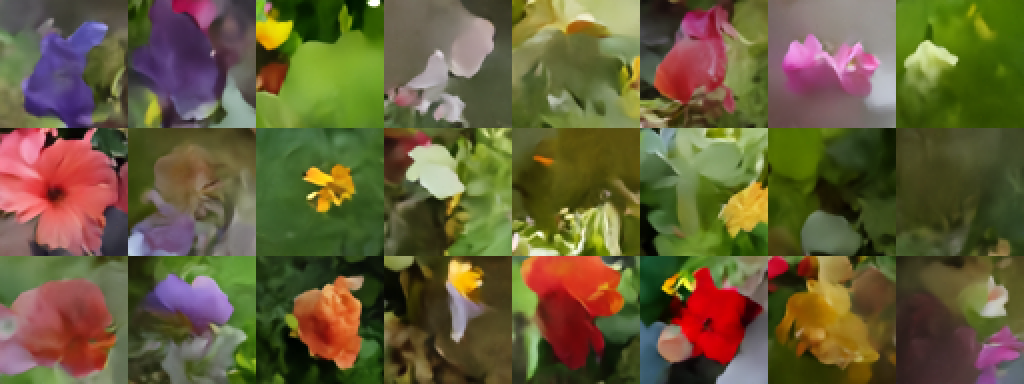}
    \\
    \includegraphics[scale=0.6]{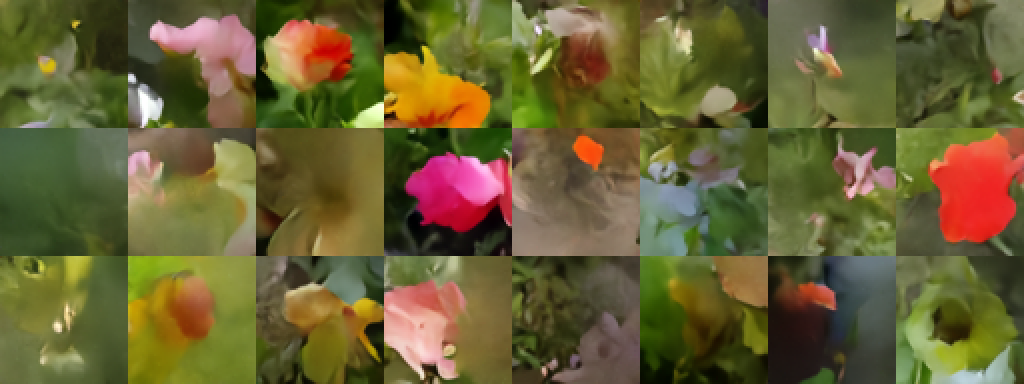}
 \caption{\textbf{Examples of images generated by the second $V_2$ (top), third $V_3$ (middle), and fourth $V_4$ (bottom) diffusion models trained with the original images and subsequent synthetic versions ($\alpha=0.5$)}}   \label{Fig:FlowersAIv234a105}
\end{figure}

\begin{figure}
    \centering
    \includegraphics[scale=0.6]{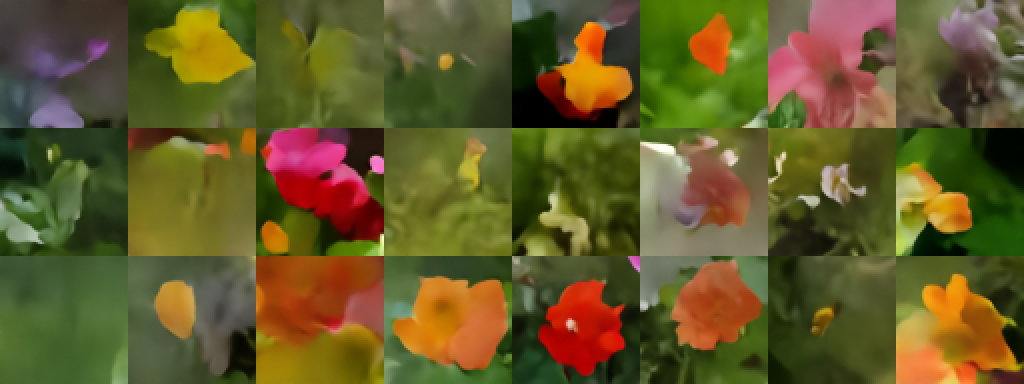}
    \\
    \includegraphics[scale=0.6]{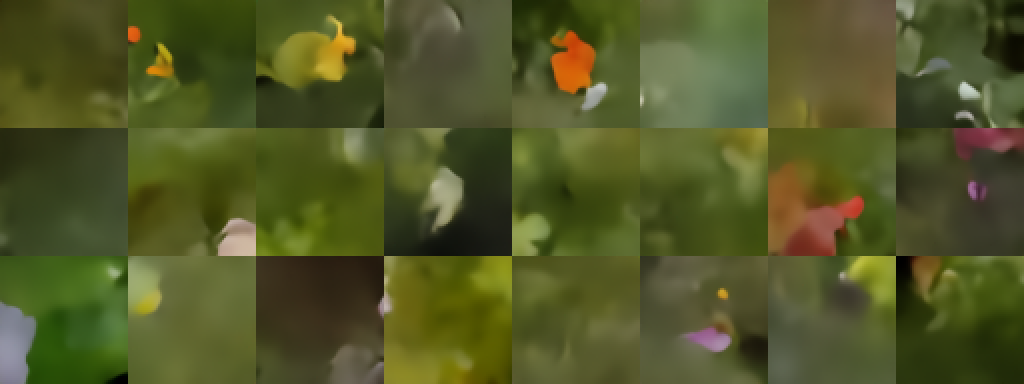}
    \\
    \includegraphics[scale=0.6]{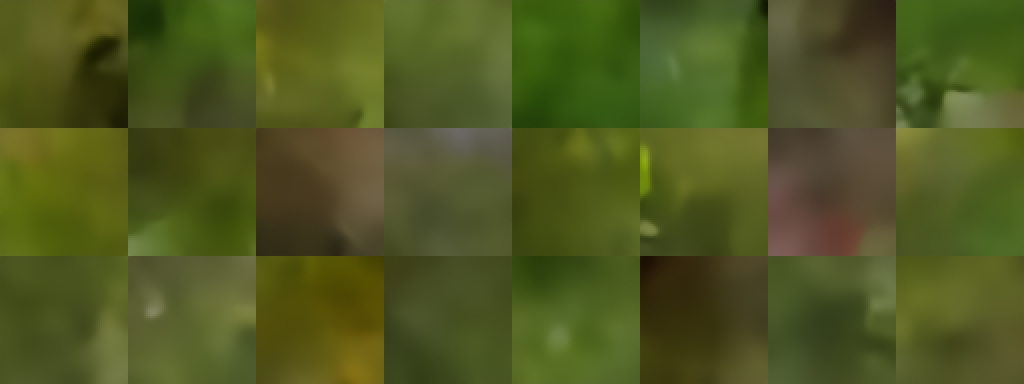}
    \caption{\textbf{Examples of images generated by the second $V_2$ (top), third $V_3$ (middle), and fourth $V_4$ (bottom) diffusion models trained with the original images and subsequent synthetic versions ($\alpha=2$)}}
    \label{Fig:FlowersAIv234a2}
\end{figure}

To see how the number of AI-generated elements impacts the quality of the diffusion models, in a second experiment we run the simulations with $\alpha= 0.5$ (so training with 2/3 real or previously generated flowers and 1/3 generated with the latest AI model on each iteration) and with $\alpha= 2$ (so adding double AI generated images to the training set on each iteration). The results are shown in Figures \ref{Fig:FlowersAIv234a105} and \ref{Fig:FlowersAIv234a2}, respectively. When $\alpha= 0.5$, the overall degradation is significantly lower but the images generated on the fourth iteration still have a much lower quality than those in the first iteration, as expected. Instead when $\alpha= 2$, the degradation is higher. This seems to indicate that the more AI-generated content added to the training data set, the lower the quality of subsequent generation of images.

\begin{figure}
    \centering
    \includegraphics[scale=0.6]{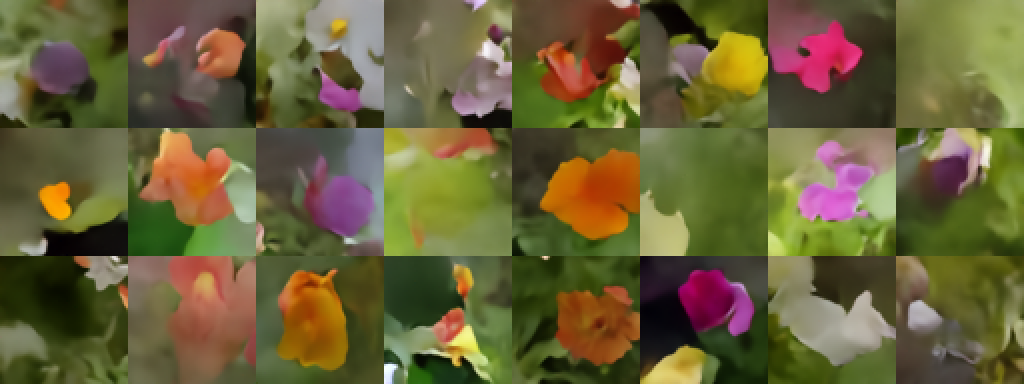}
    \includegraphics[scale=0.6]{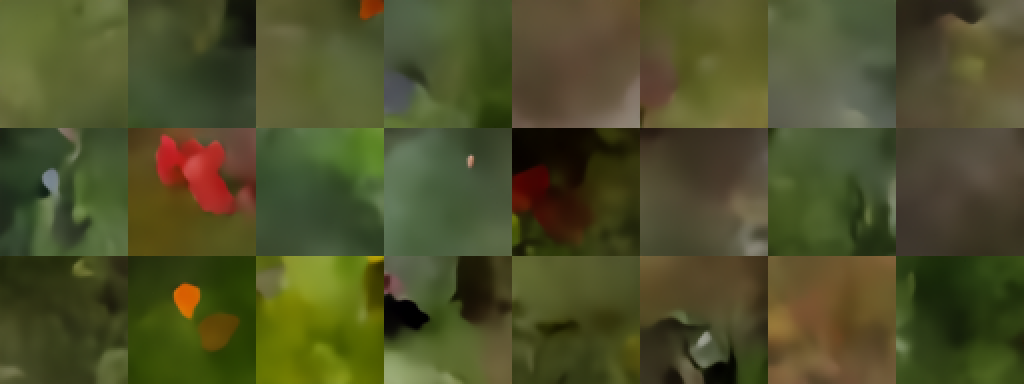}
    \includegraphics[scale=0.6]{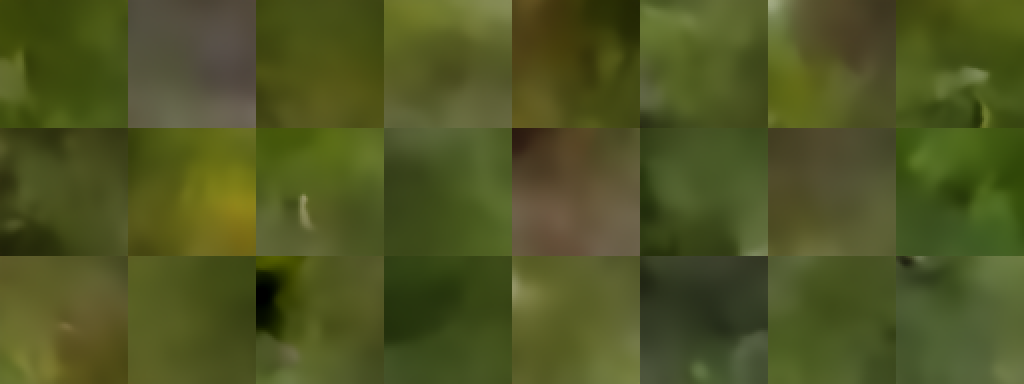}
    %\includegraphics[scale=0.6]%{./Figures/FlowersAIv5a1e.png}
    %\caption{\textbf{Examples of images generated by the second (top), third (middle), and fourth (bottom)  %diffusion models trained with the original images and those generated by previous diffusion models with %$\alpha=1$}}
   \caption{\textbf{Examples of images generated by the second $V_2$ (top), third $V_3$ (middle), and fourth $V_4$ (bottom) diffusion models trained with the original images and subsequent synthetic versions ($\alpha=1$), but doubling the number of epochs}}
    \label{Fig:FlowersAIv234a1e}
\end{figure}

Finally, we run the first experiment with $\alpha= 1$ and try to mitigate the degradation by doubling the number of epochs used for training, that is, 80 epochs. The results are shown in Figure~\ref{Fig:FlowersAIv234a1e}. We observe that even when increasing the training effort, degeneration is still present and the model starts generating noise after a few iterations.

%\section{Discussion}
%\label{sec:discussion}

%In this section, we discuss the limitations of our initial experiments and propose how to extend them. We also compare the evolution of generative AI and that of natural intelligence to understand the similarities and differences.

\section{Limitations of the experiments and future work}

The experiments in the previous section have several limitations. For example, the initial data set is small and the diffusion model is simple. Therefore, it would be interesting to run the same experiments on larger and more varied data sets to understand if degeneration occurs regardless of data set size and the nature of the images. In addition, to pinpoint the role that model complexity plays in the degeneration, it would be useful to experiment with more complex generative models.

Another interesting experiment would be to use several generative AI models to add elements to the training data set at each iteration to see if having several AI models reduces or eliminates degeneration. Additionally, each of those models could be trained on all the data set, or on different parts of it--for example, by applying mechanisms that filter which data points should be included in the next iterations, based on metrics that measure data quality or diversity.  

The main challenge is that running some of those experiments requires a large amount of computing resources that is beyond the capabilities of most university research groups.

It is also of interest to explore the evolution when AI generated content is concentrated on a type or class of images as it may happen on the Internet. Would this introduce or augment biases in newer AI generation tools? This can be explored by generating images of only one or a few classes and add them for training in the next generation. This may lead to more subtle effects than the degradation that we have observed in our simple experiments, that may have serious consequences for the fairness of these models.

\section{Conclusion}
\label{sec:Conclusion}

In this paper, we have considered the evolution of generative AI models when newer versions are trained with a mixture of real and AI-generated data. The experimental results on a simple image diffusion generative AI-model show that in each new version, the quality of the images is worse, leading to a degeneration of the AI-model's performance with each new version. Although the model and data set that we considered are simple, the results show that generative AI can suffer degradation. This suggests that more subtle effects may appear when using more complex generative models trained on larger datasets.

%In the paper, we also discuss the differences in the evolution of generative AI versus natural intelligence and also propose further experiments to evaluate how generative AI models will evolve when using more complex AI-models and data sets. 

\bibliographystyle{unsrt}  

%\begin{thebibliography}{1}
\bibliography{DeGeAI}
%\end{thebibliography}

\end{document}